\documentclass[conference]{IEEEtran}
\IEEEoverridecommandlockouts
\usepackage[utf8]{inputenc} 

\usepackage[linesnumbered,ruled,vlined]{algorithm2e} 

\usepackage{multirow}
\usepackage{booktabs}

\renewcommand\IEEEkeywordsname{Keywords}

\usepackage{cite}
\usepackage{amsmath,amssymb,amsfonts}
\usepackage{algorithmic}
\usepackage{graphicx}
\usepackage{textcomp}
\usepackage{xcolor}
\def\BibTeX{{\rm B\kern-.05em{\sc i\kern-.025em b}\kern-.08em
    T\kern-.1667em\lower.7ex\hbox{E}\kern-.125emX}}
\begin{document}

\title{Classification of autoimmune diseases from TCR repertoires by multi-instance learning}

\author{
	Ruihao Zhang\textsuperscript{1}, 
	Mao Chen\textsuperscript{2}, 
	Fei Ye\textsuperscript{3}, 
	Dandan Meng\textsuperscript{3}, 
	Yixuan Huang\textsuperscript{4}, 
	and Xiao Liu\textsuperscript{*}\\
	
	\textsuperscript{1}Tsinghua Shenzhen International Graduate School, Tsinghua University, Shenzhen, China\\
	
	*Corresponding author. Email: liuxiao@sz.tsinghua.edu.cn
}

\maketitle

\begin{abstract}
T cell receptor (TCR) repertoires encode critical immunological signatures for autoimmune diseases, yet their clinical application remains limited by sequence sparsity and low witness rates. We developed EAMil, a multi-instance deep learning framework that leverages TCR sequencing data to diagnose systemic lupus erythematosus (SLE) and rheumatoid arthritis (RA) with exceptional accuracy. By integrating PrimeSeq feature extraction with ESMonehot encoding and enhanced gate attention mechanisms, our model achieved state-of-the-art performance with AUCs of 98.95\% for SLE and 97.76\% for RA. EAMIL successfully identified disease-associated genes with over 90\% concordance with established differential analyses and effectively distinguished disease-specific TCR genes. The model demonstrated robustness in classifying multiple disease categories, utilizing the SLEDAI score to stratify SLE patients by disease severity as well as to diagnose the site of damage in SLE patients, and effectively controlling for confounding factors such as age and gender. This interpretable framework for immune receptor analysis provides new insights for autoimmune disease detection and classification with broad potential clinical applications across immune-mediated conditions.
\end{abstract}


\section{Introduction}

Autoimmune diseases result from dysregulated immune tolerance, where the immune system mistakenly targets self-tissues, triggering antibody production and cellular immune responses against endogenous components \cite{b1}. Their rising prevalence, particularly in industrialized nations, poses critical public health challenges \cite{b2,b3}. Systemic lupus erythematosus (SLE) and rheumatoid arthritis (RA) are prominent examples, characterized by multisystem organ dysfunction and chronic joint inflammation\cite{b4,b5,b6,b7}. Although early diagnosis and intervention are crucial to prevent irreversible organ damage, current diagnostic methods are complex, involving clinical, radiological, and autoantibody assessments \cite{b8,b9}.\par

T cells play a central role in autoimmune pathogenesis, with autoreactive populations implicated in SLE and RA \cite{b10}. The T cell receptor (TCR) beta chain (TRB), especially its complementarity-determining region 3 (CDR3), provides critical insights into disease mechanisms due to its diversity from V(D)J recombination \cite{b10}. High-throughput sequencing has revealed significant TCR repertoire alterations in autoimmune patients, correlating with disease progression and offering diagnostic biomarkers \cite{b11,b12,b13}. However, challenges such as low witness rates (WR), vast sequencing data, and weak-labeling paradigms—where TCR repertoires are annotated only by patient-level disease status—necessitate advanced computational approaches \cite{b14,b15,b16}.\par

Recent advancements in Transformer architectures and large language models (LLMs), such as ESM2, have enabled advanced sequence modeling by capturing complex relationships and enabling cross-domain generalization \cite{b19,b20}. Leveraging ESM2 for CDR3 sequence encoding and integrating V-gene information, we developed the ESMonehot module for comprehensive TCR feature characterization.\par

To address the unique challenges of TCR repertoire analysis, we propose EAMIL, a multimodal multi-instance learning (MIL) framework that integrates ESMonehot for feature extraction and PrimeSeq for efficient high-frequency sequence selection. EAMIL employs a multimodal fusion module to combine V-gene and CDR3 sequence information, while its attention mechanism identifies disease-associated sequences, enhancing diagnostic accuracy and interpretability. By leveraging MIL for weak-label learning, EAMIL aligns with the multimodal nature of autoimmune pathogenesis \cite{b21}.\par

We evaluated EAMIL using TCR sequencing data from SLE, RA, and healthy individuals, demonstrating state-of-the-art performance compared to DeepTCR and DeepTAPE \cite{b16,b18}. Beyond classification accuracy, EAMIL differentiated SLE populations with varying Disease Activity Index (SLEDAI) scores \cite{b27} and achieved breakthroughs in organ-damage diagnosis, highlighting its clinical utility. Main contributions include:

\begin{itemize}
\item EAMIL pioneers a multimodal MIL framework that integrates sample-level and instance-level optimization, leveraging PrimeSeq for high frequency sequence selection and ESMonehot for feature extraction. This approach effectively addresses challenges in TCR repertoire scale, witness rates, and weak-labeling scenarios.
\item Gating and spatial attention modules enable precise identification of disease-associated sequences and genes, facilitating holistic and local feature learning. These mechanisms enhance performance and interpretability, even with limited sample-level annotations.
\item EAMIL achieves state-of-the-art performance in SLE and RA classification, SLEDAI score stratification, and marks the first effective solution for diagnosing organ-specific disease damage in SLE, demonstrating its transformative potential for clinical diagnostics.
\end{itemize}

\section{Related Work}
\subsection{Multiple Instance Learning}
The autoimmune disease prediction task was formulated within a multiple instance learning (MIL) framework. The experimental dataset comprises \(N\) independent samples, denoted as \(\{S_1, S_2, \dots, S_n\}\), where each sample \(S_i\) contains \(M_i\) sequences represented as \(\{S_i^1, S_i^2, \dots, S_i^{M_i}\}\). Notably, sequence count \(M_i\) varies between samples. Each sample \(S_i\) corresponds to a diagnostic label \(L_i\), collectively forming the label set \(\{L_1, L_2, \dots, L_n\}\), where \(L\) takes values in \(\{0, \dots, C\}\), with \(C\) representing the total class count. For binary classification between autoimmune disease and healthy populations, \(C = 2\). The objective is to construct a prediction function \(F\) such that for any test sample \(S\), the function \(Y = F(S)\) predicts its corresponding disease state \(L\). 
\subsection{Disease Classification Fundation Models}
Previous computational approaches have shown limited success in addressing key challenges in TCR-based autoimmune diagnostics. Maxim E et al. utilized ensemble learning to integrate BCR and TCR data but achieved suboptimal performance on TCR data alone \cite{b17}. John-William demonstrated deep learning's potential for antigen-specific TCR classification across datasets but relied on coarse sequence handling and weak label management \cite{b16}. Tongfei et al. combined CNN and Bi-LSTM architectures for autoimmune diagnosis, yet their dictionary mapping approach struggled with large-scale CDR3 analysis, and model interpretability was insufficient \cite{b18}. The field lacks standalone deep learning frameworks capable of extracting disease-relevant sequences and gene features from weakly labeled TCR repertoires while maintaining interpretability. Furthermore, no models have achieved significant advancements in early organ damage recognition.

\section{Methods and Materials}
\subsection{Datasets}
This study leveraged a high-throughput TCR sequencing dataset of 1,522 peripheral blood samples \cite{b11}, including 877 SLE patients, 206 RA patients, and 439 healthy controls. The dataset, characterized through TCR rearrangement amplification and deep sequencing, was used for binary classification (SLE vs. Control, RA vs. Control) and multiclass classification across all three phenotypes. An 80:20 train-test split was applied with stratified sampling.

\subsection{Hyperparameters}
All experiments were conducted on an NVIDIA RTX4090 GPU to ensure computational reproducibility and consistency. Model optimization employed the Adam optimizer with a learning rate of 0.0002 and weight decay of 0.00001, balancing learning capacity and generalization. The random seed was fixed at 2024 for reproducibility. Overfitting was prevented using dropout (ratio 0.1) to enhance generalization performance on unseen data.


\begin{figure*}[!t]
	\centering
	\includegraphics[width=\textwidth]{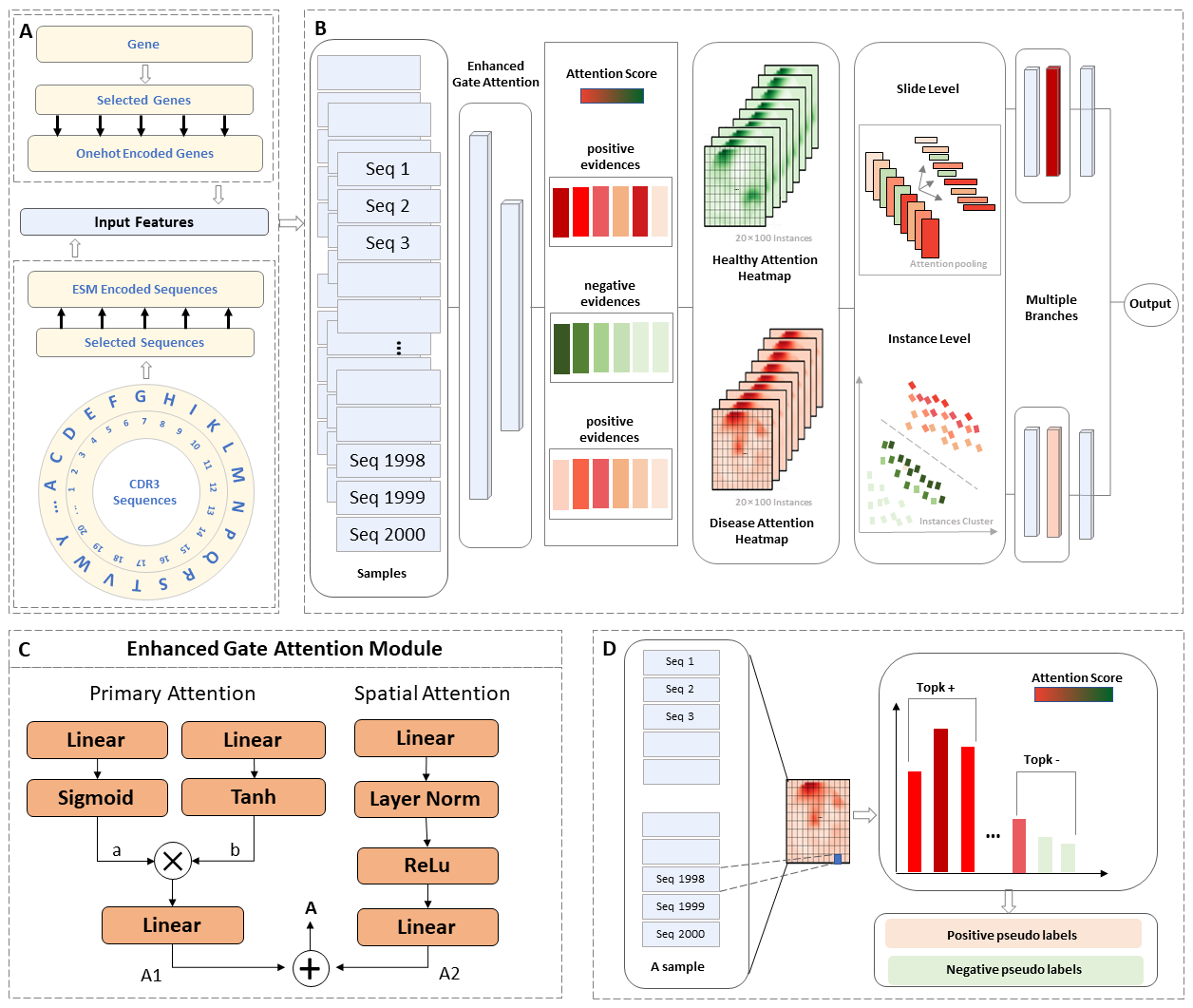}
	\caption{\textbf{Framework of the EAMIL model.} 
		(A) Feature extraction: ESMonehot encodes PrimeSeq-selected key sequences into high-dimensional vectors. 
		(B) Multi-instance learning: Enhanced gate attention mechanism with pooling and clustering strategy uses bag-level labels as pseudo-labels to optimize sequence and sample-level feature learning. 
		(C) Dual attention: Gated and spatial attention mechanisms enable multi-level modeling of global and local features. 
		(D) Attention-based labeling: Top-k sequences receive pseudo-labels based on attention scores from the enhanced gate attention mechanism.}
	\label{fig:first_full_page}
\end{figure*}

\subsection{Model Architecture}

We present EAMIL, a deep learning framework tailored for TCR sequence analysis. Leveraging the PrimeSeq strategy, EAMIL integrates ESMonehot feature extraction, combining a pretrained ESM encoder for CDR3 sequences and one-hot encoding for TCR genes to generate high-dimensional representations. Through multi-instance learning with an enhanced gated attention mechanism and a clustering strategy using bag-level pseudo-labels, EAMIL effectively learns features at both sequence and sample levels, optimizing training and improving classification performance, as shown in Fig.~\ref{fig:first_full_page}.

\subsubsection{Feature Extraction Module}

TCR sequencing generates large datasets with significant background noise, posing challenges in identifying rare disease-associated signals. The PrimeSeq strategy addresses this by leveraging frequency information to extract the top 2,000 most frequent sequences from each sample, reducing encoding requirements and mitigating issues such as low witness rates (WR) and vast TCR sequence diversity.\par

To comprehensively profile TCR repertoires, we developed a multimodal fusion encoding module that integrates CDR3 sequences with gene information. Using the PrimeSeq-selected sequences, the pretrained ESM model encodes CDR3 sequences, while genes are represented via one-hot encoding. These encodings are concatenated into a fusion feature matrix, followed by pooling to generate sequence-level features. This combination of protein language pretraining and multimodal fusion enriches TCR receptor representations, enhancing disease prediction and classification.\par

\subsubsection{Enhanced Gate Attention Module}
The Enhanced Gated Attention Module is a dual-pathway attention mechanism designed to optimize feature extraction by integrating global contextual dependencies and spatial sensitivity. This module addresses the limitations of conventional attention mechanisms in simultaneously capturing broad contextual patterns and localized feature variations, enabling the extraction of enriched representations critical for downstream tasks such as classification and disease prediction. By combining a primary gated attention mechanism with an auxiliary spatial attention branch, the module achieves a synergistic feature representation that enhances model performance in complex whole slide image analysis.

The primary attention branch operates on input features $x \in \mathbb{R}^{N \times L}$, where $N$ denotes the number of patches and $L$ represents the feature dimensionality. This branch applies two parallel transformations. The first one is non-linear activation using Tanh:
\begin{equation}
	a = \tanh(W_a x + b_a), \quad W_a \in \mathbb{R}^{D \times L}, \quad b_a \in \mathbb{R}^D
\end{equation}
Then, it takes Sigmoid-based gating:
\begin{equation}
	b = \sigma(W_b x + b_b), \quad W_b \in \mathbb{R}^{D \times L}, \quad b_b \in \mathbb{R}^D
\end{equation}
The outputs of these two transformations are multiplied element-wise to generate the gated attention weights:
\begin{equation}
	A_1 = W_c \cdot (a \odot b) + b_c, \quad W_c \in \mathbb{R}^{n_{classes} \times D}, \quad b_c \in \mathbb{R}^{n_{classes}}
\end{equation}
where $n_{classes}$ is the number of disease classes. This gated attention mechanism ensures that the module selectively amplifies informative patterns while suppressing noise, capturing global contextual dependencies.

To complement the global attention from the primary branch, the module incorporates a spatial attention branch designed to capture localized feature patterns. This branch applies a series of linear transformations, normalization, and non-linear activation to the input $x$, defined as:
\begin{equation}
	A_2 = W_s^{(2)} \cdot \text{ReLU}(\text{LayerNorm}(W_s^{(1)} x + b_s^{(1)})) + b_s^{(2)}
\end{equation}
where $W_s^{(1)} \in \mathbb{R}^{(D/2) \times L}$, $W_s^{(2)} \in \mathbb{R}^{n_{classes} \times (D/2)}$, and $b_s^{(1)}, b_s^{(2)}$ are bias terms. The spatial attention branch allows the module to focus on fine-grained details, making it particularly effective in tasks where localized variation is important.

The outputs from the two branches are combined to form the final attention score:
\begin{equation}
	A = A_1 + \lambda A_2
\end{equation}
where $\lambda = 0.5$ is a scaling factor that balances the contributions of global and spatial attention. This fused attention score is subsequently used to refine feature representations for downstream tasks, as shown in Fig.~\ref{fig:first_full_page}C.\par
	

\subsubsection{Multi-Instance Learning Module}
\begin{itemize}
	\item Attention-Pooling Mechanism: Each sample $z \in \mathbb{R}^{N \times L}$ is represented as a feature matrix consisting of $N$ feature vectors, where $L$ is the dimensionality of the feature space. These feature vectors are derived from prior encoding steps, and each corresponds to an instance within the sample. The Multiple Instance Learning (MIL) setting assumes that only sample-level labels are available, while individual instance labels remain unknown.
	
	Using the attention scores $A$ computed by the Enhanced Gate Attention Module (Equation 5), we normalize the scores across all instances within a sample using the softmax function:
	\begin{equation}
		a_k = \frac{\exp(A_k)}{\sum_{i=1}^{N} \exp(A_i)}
	\end{equation}
	where $a_k$ represents the normalized attention weight for instance $k$.
	
	\item Sample-Level Representation: The instance-level features are aggregated into a sample-level representation $M \in \mathbb{R}^L$ using the attention weights:
	\begin{equation}
		M = \sum_{k=1}^N a_k h_k
	\end{equation}
	This weighted aggregation ensures that the most informative instances contribute more significantly to the final sample representation, improving the model's ability to classify complex and noisy data.
	
	\item Classification Heads: The aggregated feature representation $M$ is passed through a disease classifier for the main task classification:
	\begin{equation}
		y = \text{softmax}(W_c M + b_c)
	\end{equation}
	where $W_c \in \mathbb{R}^{n_{\text{classes}} \times L}$ and $b_c \in \mathbb{R}^{n_{\text{classes}}}$ are trainable weights and biases. Then, $M$ is passed through the location classifier to predict spatial or positional labels:
	\begin{equation}
		l = \text{softmax}(W_l M + b_l)
	\end{equation}
	where $W_l \in \mathbb{R}^{n_{\text{locations}} \times L}$ and $b_l \in \mathbb{R}^{n_{\text{locations}}}$ are trainable weights and biases.\par

\end{itemize}


\subsubsection{Training Strategy}
\begin{itemize}
	\item Instance-Level Supervision: To bridge the gap between instance-level representations and sample-level labels, the model generates pseudo-labels for the top-$k$ most informative instances based on their attention scores. These pseudo-labels are derived by combining true sample-level labels with the predicted attention distribution, as shown in Fig.~\ref{fig:first_full_page}D. For the top-$k$ instances, the pseudo-labels are used to compute an instance-level loss:
	\begin{equation}
		L_p = \frac{1}{N_P} \sum_{i=1}^{N_P} L_{\text{SVM}}(\hat{y}_i, y_i)
	\end{equation}
	where $L_{\text{SVM}}$ denotes the smoothed top-1 SVM loss, $\hat{y}_i$ represents the predicted label for instance $i$, $y_i$ is the corresponding pseudo-label, and $N_P$ is the number of pseudo-labeled instances.
	\item Sample-Level Supervision: The primary classification loss compares the predicted sample-level labels $\hat{y}$ with the ground-truth labels $y$ using a cross-entropy loss:
	\begin{equation}
		L_s = L_{\text{CE}}(\hat{y}, y)
	\end{equation}
	
	The total loss $L_{\text{total}}$ is computed as a weighted combination of the sample-level and instance-level losses:
	\begin{equation}
		L_{\text{total}} = c_1 L_s + c_2 L_p
	\end{equation}
	where $c_1$ and $c_2$ are hyperparameters that determine the balance between sample-level and instance-level supervision, with $c_1 + c_2 = 1$.\par

\end{itemize}

\section{RESULTS}

EAMIL integrates feature extraction, an enhanced gate attention module, and multi-instance deep learning to analyze complex immunogenomic data. Using high-throughput sequencing data from 1,522 individuals (877 SLE, 206 RA, 439 controls), EAMIL demonstrates state-of-the-art performance in binary and multi-classification tasks, including disease diagnosis, stratification by SLEDAI scores, and analysis of factors like age and sex. Key findings are validated through interpretability analysis, feature space visualization, and diagnostic evaluations, showcasing EAMIL's robustness and potential for autoimmune disease diagnostics.

\subsection{Disease Prediction with Transfer-Learning Comparison}
We evaluated the ESMonehot encoding module alongside various transfer learning models for SLE prediction, including attention-based frameworks (ABMIL, G-ABMIL, RRT, TransMIL), clustering-based methods (CLAM-SB, CLAM-MB), and other MIL algorithms (MeanMIL, MaxMIL, WIKG), adapted from pathology image analysis \cite{b22,b23,b24,b25,b26}. By replacing their original coding modules with ESMonehot and transfering architectures to autoimmune classification, we demonstrated that EAMIL, combining ESMonehot with enhanced gate attention, achieved SOTA performance in SLE and RA binary classification tasks, as shown in Table I.\par

\begin{table}[htbp]
	\centering
	\caption{Performance Comparison Across Methods for SLE and RA}
	\resizebox{\linewidth}{!}{ 
		\begin{tabular}{p{1.3cm}p{1.6cm}p{1.3cm}p{1.3cm}p{1.3cm}p{1.3cm}p{1.3cm}}
			\toprule
			\textbf{Disease} & \textbf{Methods} & \textbf{ACC (\%)} & \textbf{AUC (\%)} & \textbf{Pre (\%)} & \textbf{Recall (\%)} & \textbf{F1 (\%)} \\ 
			\midrule
			\multirow{10}{*}{SLE} 
			& ABMIL & 90.31 & 95.69 & 89.65 & 88.54 & 88.99 \\ 
			& G-ABMIL & 93.44 & 97.93 & 92.89 & 92.36 & 92.60 \\ 
			& RRT & 72.62 & 79.43 & 69.29 & 69.53 & 69.32 \\ 
			& CLAM-SB & 90.47 & 95.78 & 89.87 & 88.65 & 89.15 \\ 
			& CLAM-MB & 93.67 & 98.16 & 93.33 & 92.42 & 92.82 \\ 
			& MeanMIL & 93.29 & 97.95 & 92.94 & 91.90 & 92.37 \\ 
			& MaxMIL & 74.91 & 80.66 & 72.36 & 69.04 & 70.00 \\ 
			& TransMIL & 89.02 & 93.95 & 87.74 & 87.56 & 87.62 \\ 
			& WIKG & 93.36 & 97.55 & 92.79 & 92.24 & 92.50 \\ 
			& \textbf{EAMIL} & \textbf{95.12} & \textbf{98.95} & \textbf{94.59} & \textbf{94.47} & \textbf{94.51} \\ 
			\midrule
			\multirow{10}{*}{RA} 
			& ABMIL & 88.06 & 94.42 & 86.58 & 86.59 & 86.43 \\ 
			& G-ABMIL & 90.23 & 95.98 & 89.38 & 87.93 & 88.57 \\ 
			& RRT & 71.32 & 70.83 & 66.72 & 64.73 & 65.16 \\ 
			& CLAM-SB & 89.77 & 95.53 & 88.65 & 88.11 & 88.25 \\ 
			& CLAM-MB & 88.99 & 95.50 & 88.20 & 86.63 & 87.19 \\ 
			& MeanMIL & 83.72 & 92.69 & 81.98 & 79.81 & 80.67 \\ 
			& MaxMIL & 73.95 & 77.28 & 72.29 & 64.24 & 64.87 \\ 
			& TransMIL & 79.38 & 87.92 & 76.81 & 75.18 & 75.74 \\ 
			& WIKG & 83.72 & 91.58 & 81.39 & 82.39 & 81.62 \\ 
			& \textbf{EAMIL} & \textbf{92.71} & \textbf{97.76} & \textbf{91.62} & \textbf{91.70} & \textbf{91.62} \\ 
			\bottomrule
			\multicolumn{7}{l}{\footnotesize All results are from 5-fold cross-validation with std. dev. $< 0.1$.}
		\end{tabular}
	}
	\label{tab:performance}
\end{table}

The multi-branching CLAM-MB consistently outperformed single-branching CLAM-SB using V gene and CDR3 features, while EAMIL outperformed both. This reflects EAMIL's ability to capture subtle TCR rearrangement differences in autoimmune patients, facilitated by its enhanced gate attention and multi-branching architecture. Results validated ESMonehot's encoding efficiency and PrimeSeq's effectiveness, with five-fold cross-validation showing rapid model convergence (10 epochs) and stable performance after 20 epochs.\par
We further validated EAMIL on RA datasets, maintaining the same encoding and feature selection. EAMIL consistently outperformed other models across all metrics, confirming its adaptability and robustness in autoimmune disease prediction. Additionally, G-ABMIL with ESMonehot achieved impressive results, further highlighting the reliability of our encoding methodology and MIL frameworks.

\subsection{Comparison with Existing Methods}
We compared EAMIL with state-of-the-art algorithms, DeepTCR and DeepTAPE, using datasets from SLE, RA, and healthy individuals \cite{b16,b18}. DeepTCR models complex TCR sequences for repertoire classification, while DeepTAPE combines CNN and LSTM modules for unified classification. Using identical V gene and CDR3 sequence features and consistent learning parameters, we conducted binary classification tasks (SLE vs. Control, RA vs. Control) with five-fold cross-validation and 50 training epochs.\par
\begin{figure}[htbp]
	\centering
	\includegraphics[width=0.8\columnwidth]{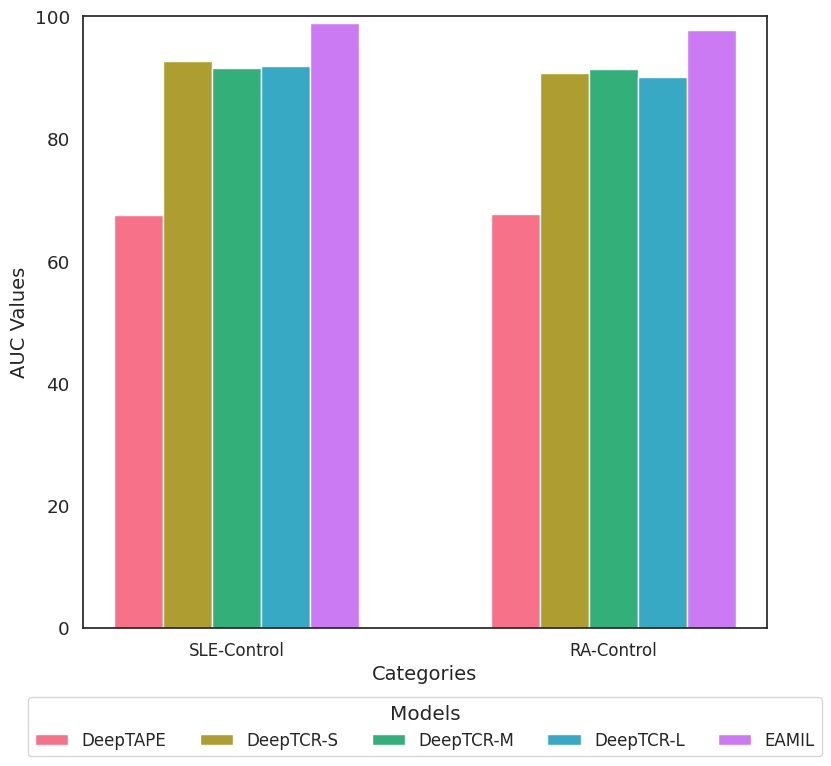} 
	\caption{\textbf{Comparison of results with existing deep learning methods.} The figure illustrates the comparative performance of our model against DeepTCR (in small, medium, and large settings) and DeepTAPE on two disease cases using the AUC metric.}
	\label{fig:second_inline}
\end{figure}

EAMIL consistently outperformed DeepTCR and DeepTAPE in AUC metrics across both disease classification tasks, as shown in Fig.~\ref{fig:second_inline}. These results validate EAMIL's superior performance in TCR sequence classification and its potential for analyzing complex immunogenomic data.


\subsection{Significant Genes Identification}
\begin{figure*}[!t]
	\centering
	\includegraphics[width=\textwidth]{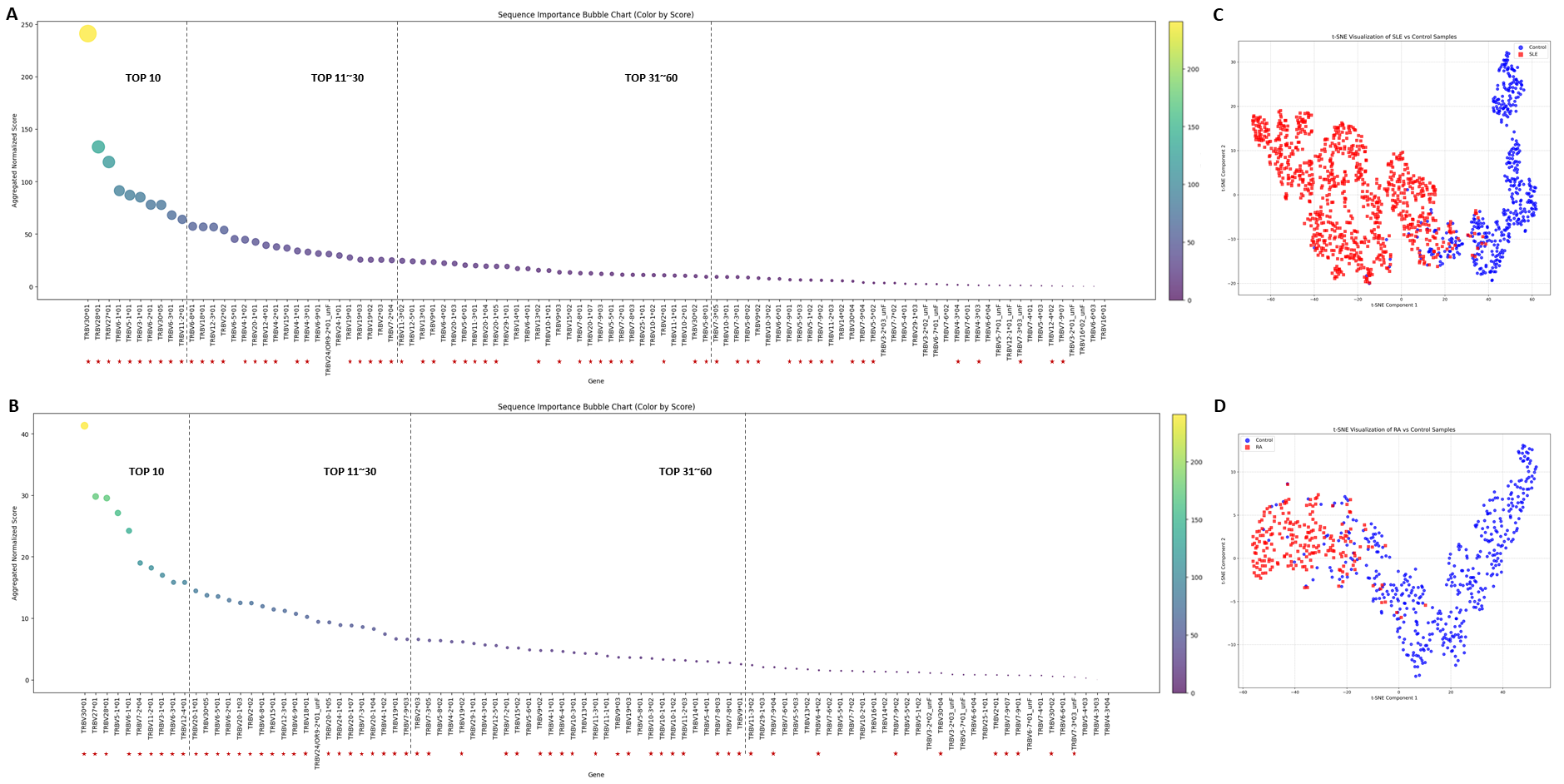} 
	\caption{\textbf{Visual analysis of significant genes and features.} To identify significant genes in SLE and rheumatoid arthritis patients, the top 10 sequences cumulatively identified by our model were counted and visualized as bubble plots. Each bubble represents a gene, with higher scores indicating greater significance. Genes that align with findings from previous studies are highlighted with red stars (A, B). Furthermore, T-SNE plots display the distribution of SLE-encoded VCDR3 features (C) and RA-encoded VCDR3 features (D) extracted from the attention module for SLE controls and RA controls, respectively.}
	\label{fig:third_full_page}
\end{figure*}

To validate EAMIL's ability to identify immunologically significant features, we analyzed attention heatmaps from randomly selected SLE and RA samples. For SLE, attention scores revealed the top 10 sequences, aligning precisely with independently identified disease-specific sequences \cite{b11}, including TRBV13 and TRBV5 gene families, which are exclusive to SLE patients (Fig.\ref{fig:third_full_page}A). Similarly, in RA, 28 of the top 30 sequences matched previously reported RA-associated sequences (Fig.\ref{fig:third_full_page}B), demonstrating EAMIL's robust capability to identify disease-relevant genes.\par
To further visualize feature separation, t-SNE plots for SLE-Control (Fig.\ref{fig:third_full_page}C) and RA-Control (Fig.\ref{fig:third_full_page}D) comparisons showed near-complete separation between disease states. These results underscore the effectiveness of the PrimeSeq strategy, ESMonehot encoding, and enhanced gate attention in feature learning and discrimination, offering valuable insights for disease analysis and therapeutic development.\par

\begin{table}[htbp]
	\centering
	\caption{Gene Validation of the Systemic Lupus Erythematosus and Rheumatoid Arthritis Prediction}
	\resizebox{\linewidth}{!}{ 
		\begin{tabular}{p{1.5cm}p{2cm}p{2.5cm}p{2.5cm}}
			\toprule
			\textbf{Disease} & \textbf{Genes} & \textbf{Relationship} & \textbf{Validation} \\ 
			\midrule
			\multirow{10}{*}{SLE} 
			& TRBV28*01 & downregulation & Direct \cite{b26} \\ 
			& TRBV27*01 & upregulation & Indirect \cite{b27} \\ 
			& TRBV6-1*01 & other activity & Direct \cite{b28} \\ 
			& TRBV5-1*01 & downregulation & Direct \cite{b26} \\ 
			& TRBV3-1*01 & downregulation & Direct \cite{b26} \\ 
			& TRBV6-2*01 & downregulation & Direct \cite{b26,b28} \\ 
			& TRBV6-3*01 & downregulation & Direct \cite{b26} \\ 
			& TRBV11-2*01 & other activity & Direct \cite{b26} \\ 
			& TRBV6-8*01 & upregulation & Indirect \cite{b29} \\ 
			& TRBV18*01 & other activity & Indirect \cite{b29} \\ 
			& TRBV2*02 & other activity & Direct \cite{b30} \\ 
			\midrule
			\multirow{10}{*}{RA} 
			& TRBV27*01 & upregulation & Indirect \cite{b27} \\ 
			& TRBV28*01 & downregulation & Direct \cite{b11} \\ 
			& TABV5-1*01 & other activity & Direct \cite{b31} \\ 
			& TRBV6-1*01 & other activity & Direct \cite{b11} \\ 
			& TRBV7-2*04 & upregulation & Direct \cite{b27} \\ 
			& TRBV11-2*01 & other activity & Indirect \cite{b32} \\ 
			& TRBV6-3*01 & other activity & Indirect \cite{b33} \\ 
			& TRBV12-4*01 & upregulation & Direct \cite{b29} \\ 
			& TRBV20-1*01 & upregulation & Direct \cite{b29} \\ 
			& TRBV6-2*01 & other activity & Direct \cite{b11} \\ 
			& TRBV2*02 & other activity & Direct \cite{b30} \\ 
			\bottomrule
		\end{tabular}
	}
	\label{tab:gene_validation}
\end{table}

The genes identified by the model as associated with the pathogenesis of systemic lupus erythematosus (SLE) and rheumatoid arthritis (RA) were validated through an extensive literature review, with the majority of top-ranked genes directly or indirectly implicated in disease mechanisms. Notably, the $TRBV28^*01$ gene exhibited a down-regulated effect in both SLE and RA, while the $TRBV27^*01$ gene demonstrated an indirect up-regulation effect in both diseases. Studies further revealed that the $TRBV6$-$1^*01$ gene is significantly enriched in SLE patients, whereas the $TRBV6$-$2^*01$ gene is markedly down-regulated. The $TRBV2^*02$ gene showed a positive correlation with the SLEDAI.\par
In RA, the $TABV5$-$1^*01$ gene was highly expressed in T cells infiltrating synovial tissues, and $TRBV11$-$2^*01$ was linked to Th17-mediated inflammation. Additionally, the $TRBV2^*02$ gene showed a significant correlation with RA disease activity, as measured by the Disease Activity Score (DAS28). These findings, as summarized in Table~\ref{tab:gene_validation}, strongly align with prior studies, underscoring the effectiveness of EAMIL’s feature extraction and enhanced attention module. The robust performance of this model provides a powerful framework for identifying disease-associated genes, with profound implications for clinical diagnosis, disease analysis, and therapeutic development.

\subsection{Multiclassification Comparative Experiments}
To evaluate EAMIL's multiclassification capabilities, we conducted a three-class prediction task (SLE, arthritis, and healthy controls) using TCR rearrangement data from 1522 individuals. EAMIL outperformed G-ABMIL with a 3.4\% higher AUC, 6.99\% higher accuracy. Notably, EAMIL's F1 score greatly exceeds that of all other transfer learning models, confirming its great strength in multi-class discrimination, as shown in Table~\ref{tab:multiclassification}. These results highlight EAMIL's exceptional performance in multi-class discrimination, critical for clinical disease diagnosis and differential analysis.

\begin{table}[htbp]
	\centering
	\caption{Results of Multiclassification}
	\resizebox{\linewidth}{!}{ 
		\begin{tabular}{p{2.5cm}p{1.3cm}p{1.3cm}p{1.3cm}p{1.3cm}p{1.3cm}}
			\toprule
			\textbf{Methods} & \textbf{ACC (\%)} & \textbf{AUC (\%)} & \textbf{Pre (\%)} & \textbf{Recall (\%)} & \textbf{F1 (\%)} \\ 
			\midrule
			ABMIL & 80.23 & 92.01 & 72.04 & 63.21 & 61.97 \\ 
			G-ABMIL & 80.29 & 92.66 & 70.36 & 65.45 & 63.49 \\ 
			RRT & 64.86 & 72.50 & 45.68 & 48.87 & 46.69 \\ 
			CLAM-SB & 81.54 & 93.08 & 72.10 & 65.00 & 64.14 \\ 
			CLAM-MB & 81.41 & 93.20 & 71.06 & 66.79 & 67.02 \\ 
			MeanMIL & 80.69 & 92.93 & 61.15 & 61.51 & 57.93 \\ 
			MaxMIL & 62.10 & 70.52 & 39.89 & 43.94 & 41.21 \\ 
			TransMIL & 77.06 & 84.26 & 68.34 & 67.03 & 67.50 \\ 
			WIKG & 81.34 & 93.42 & 70.63 & 65.83 & 65.16 \\ 
			\textbf{EAMIL} & \textbf{87.28} & \textbf{96.06} & \textbf{82.27} & \textbf{79.28} & \textbf{80.39} \\ 
			\bottomrule
			\multicolumn{6}{l}{\footnotesize All results are from 5-fold cross-validation with std. dev. $< 0.1$.}
		\end{tabular}
	}
	\label{tab:multiclassification}
\end{table}

\subsection{One Versus Others Experiments}
\begin{figure}[htbp]
	\centering
	\includegraphics[width=\columnwidth]{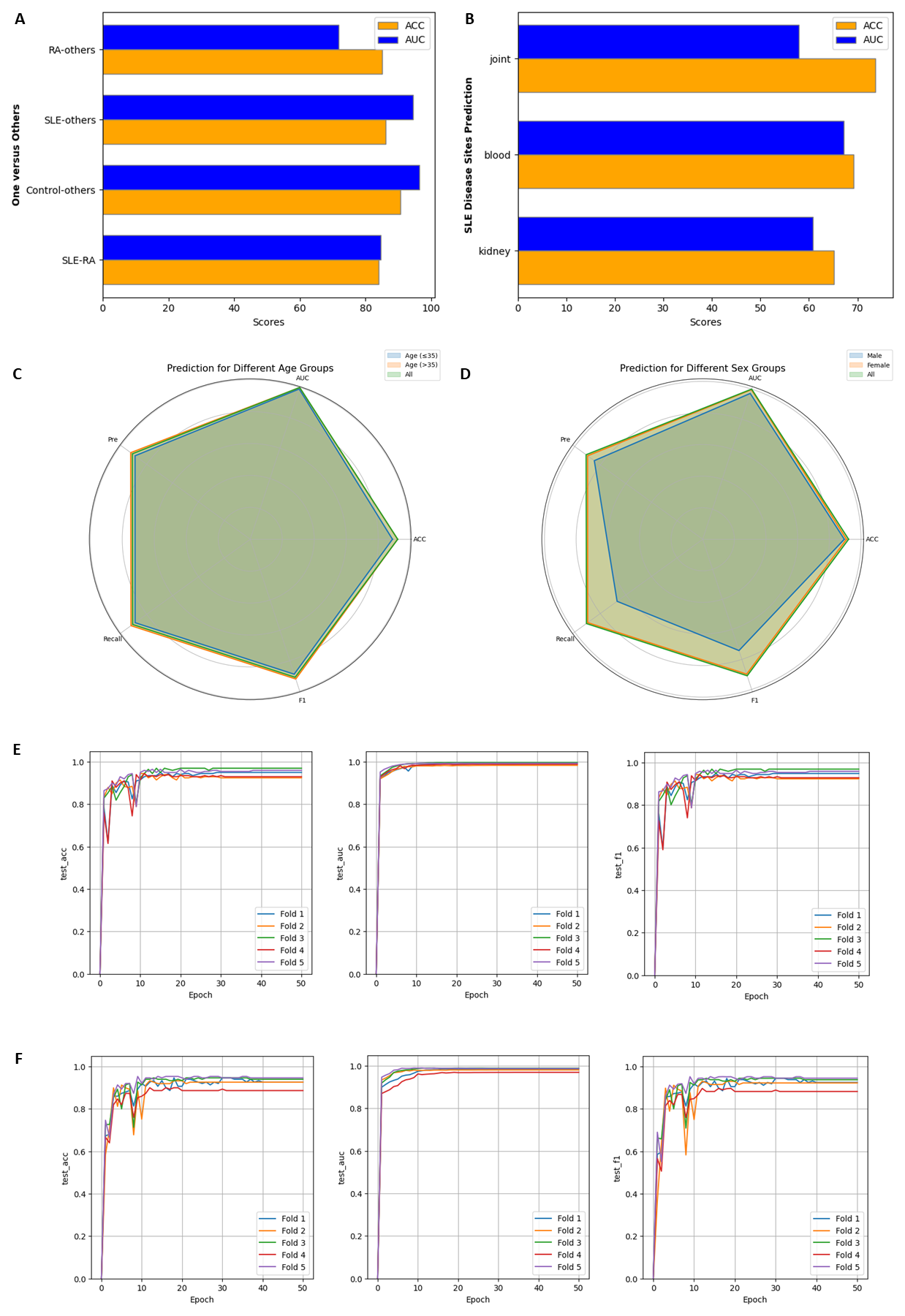} 
	\caption{\textbf{Results of analysis experiments.} (A) One-vs-others experiment results, illustrating the classification performance for SLE, RA, and Control groups. (B) Diagnostic analysis identifying damaged body parts in SLE patients, with affected areas labeled. (C) Sex-based analysis of SLE patients, highlighting differences across genders. (D) Age group-based analysis of SLE patients, showcasing variations across different age groups. Radar charts show five-fold cross-validation results, adjusted by subtracting 50\%. (E) Identification of Active SLE patients compared to healthy samples. (F) Identification of Silent SLE patients compared to healthy samples.}
	\label{fig:fourth_inline}
\end{figure}

To evaluate EAMIL's ability to identify specific disease states within a heterogeneous population, one versus others experiments were conducted on 1,522 individuals, targeting systemic lupus erythematosus (SLE) and rheumatoid arthritis (RA). EAMIL achieved strong performance in SLE detection, with an AUC of 94.53\% and an accuracy of 86.08\% (Fig.~\ref{fig:fourth_inline}A). While RA detection was less robust, likely due to sample size imbalance and feature overlap, the model performed well in distinguishing SLE from RA, with an AUC of 84.66\% and an accuracy of 83.95\%. Additionally, EAMIL excelled in separating healthy individuals from autoimmune patients, achieving an AUC of 96.28\% and an accuracy of 90.51\%. These results highlight EAMIL's potential for accurate autoimmune disease diagnosis and clinical application.

\subsection{SLE Organ-specific Damage Identification}
To further leverage our model’s exceptional diagnostic performance, we explored its potential in identifying organ damage in systemic lupus erythematosus (SLE) patients—a critical step for improving clinical management and prognosis. The study focused on classifying damage versus non-damage in the blood, kidney, and joint systems using clinical data from SLE patients, with a five-fold cross-validation approach. Notably, C3 and C4 levels, key components of the complement system and critical markers of immune function, were incorporated into the analysis, given their established role in autoimmune disease diagnostics \cite{b36,b37}.\par
By integrating sample information with feature tensor, the model demonstrated significant advancements in organ-specific damage detection, as shown in Fig.~\ref{fig:fourth_inline}B. In the blood damage classification task, it achieved an AUC of 67.18\%, markedly outperforming DeepTCR’s AUC of 55.97\%. These results underscore the model’s potential to accurately identify SLE-related organ damage and provide valuable insights for clinical diagnosis and personalized treatment strategies.
\subsection{SLEDAI Comparative Experiment}
The Systemic Lupus Erythematosus Disease Activity Index (SLEDAI) is a key clinical tool for assessing SLE activity, guiding treatment, and monitoring disease progression \cite{b25}. Using TCR repertoire data from 873 SLE patients (312 silent, 561 active) and 439 healthy controls, we evaluated EAMIL's performance in distinguishing SLE activity states.\par
EAMIL demonstrated enhanced discriminatory capacity for active SLE, attaining an F1 score of 94.61\% (2.36\% superior to silent SLE), reflecting the more pronounced immunological perturbations characteristic of active disease states (Fig.\ref{fig:fourth_inline}E, Fig.\ref{fig:fourth_inline}F). For the more clinically challenging discrimination between silent SLE and controls, EAMIL sustained robust performance with an AUC of 98.26\% and accuracy of 92.52\%.\par
While active disease states provided substantial signal for model optimization, the detection of silent SLE represents a clinically significant capability for comprehensive SLE diagnosis. These findings validate EAMIL's efficacy across the spectrum of disease activity and underscore its translational potential for early detection and precision therapeutic intervention in autoimmune pathology.

\subsection{Limited Impact of Age and Sex on SLE Disease Identification}
To assess potential demographic influences on EAMIL's SLE classification, we analyzed 873 SLE patients stratified by age and sex against 439 healthy controls. Stratification by age (median: 35 years) revealed superior classification performance for patients $\leq 35$ years, likely reflecting age-related immunological changes such as immune aging, chronic inflammation, and enhanced autoimmune responses (Fig.~\ref{fig:fourth_inline}C).\par

Sex-based analysis showed better performance for female SLE patients (n=774) versus controls compared to male patients (n=99), consistent with SLE's higher prevalence in females and estrogen's immunomodulatory effects that exacerbate autoimmune features. Reduced performance in males was also influenced by the smaller sample size (Fig.~\ref{fig:fourth_inline}D).\par

While older and female patients exhibited more distinct disease signatures, the modest differences in accuracy and AUC across demographic strata confirm EAMIL’s robust ability to learn disease-specific features with minimal impact from demographic factors.

\section{Discussion and Conclusion}
In this study, we successfully developed and validated EAMIL, a multi-instance deep learning model leveraging the PrimeSeq sequence extraction strategy for TCR sequence analysis and autoimmune disease identification. By employing ESMonehot encoding, EAMIL integrates diverse information sources, including gene information and amino acid sequences, while its enhanced attention mechanism provides interpretability by identifying disease-relevant features. This foundation enables instance clustering, pseudo-label generation, and robust autoimmune disease classification by analyzing extensive TCR sequence data and prioritizing key sequences through quantitative attention scores.\par

Experimental results demonstrate EAMIL's exceptional performance across multiple tasks, particularly in identifying systemic lupus erythematosus (SLE) and rheumatoid arthritis (RA). The model achieved state-of-the-art metrics in SLE prediction, excelling in AUC and accuracy, while effectively distinguishing SLE patients stratified by SLEDAI scores from healthy controls. EAMIL also showed promise in diagnosing organ damage in SLE patients and maintained strong performance despite demographic confounders such as age and sex. Moreover, the model successfully identified disease-relevant genes, enabling accurate differentiation between distinct autoimmune conditions—findings with significant clinical implications for diagnosis and therapeutic development.\par

Future research will focus on expanding EAMIL's capabilities by incorporating additional autoimmune disease datasets to enhance performance and generalizability. Integrating clinical data for more comprehensive disease characterization and extending applications to other autoimmune diseases represent promising directions. Cross-disciplinary approaches will further facilitate the identification of disease-associated receptors, paving the way for new diagnostic and therapeutic advancements.\par

As deep learning technologies and biomedical data resources continue to evolve, EAMIL is poised to make significant contributions to precision medicine. With ongoing algorithm optimization and model refinement, EAMIL has the potential to become a vital diagnostic tool for the prediction and classification of autoimmune diseases.


\end{document}